\documentclass{article}

\usepackage[preprint]{neurips_2026}

\usepackage[utf8]{inputenc} 
\usepackage[T1]{fontenc}    
\usepackage{hyperref}       
\usepackage{url}            
\usepackage{booktabs}       
\usepackage{amsfonts}       
\usepackage{nicefrac}       
\usepackage{microtype}      
\usepackage{graphicx}       
\usepackage{flafter}        
\usepackage{placeins}       
\usepackage{caption}

\usepackage{amsmath}
\usepackage{amssymb}
\usepackage[table]{xcolor}
\definecolor{dashhighlight}{RGB}{255,250,230}

\title{DASH: Fast Differentiable Architecture Search for Hybrid Attention in Minutes on a Single GPU}

\author{%
  Weizhe Chen \quad
  Miao Zhang$^\dagger$ \quad
  Junpeng Jiang \quad 
  Yaping Li \quad 
  Weili Guan \quad 
  Liqiang Nie \\
  Harbin Institute of Technology (Shenzhen) \\
  \texttt{weizhechen@stu.hit.edu.cn} \\
  \texttt{zhangmiao@hit.edu.cn} \\
  $^{\dagger}$Corresponding author.
}

\begin{document}

\maketitle

\begin{abstract}
Hybrid attention architectures are becoming an increasingly important
paradigm for improving LLM inference efficiency while preserving model
quality,
making hybrid architecture design a central problem.
Existing designs often rely on manual empirical rules or proxy-based
selector signals for layer-wise operator allocation.
Recent NAS-style systems such as Jet-Nemotron demonstrate the promise of
automated hybrid architecture search.
However,
Jet-Nemotron's PostNAS search stages alone use 200B tokens,
making such search pipelines difficult to use as routine methods for
hybrid architecture design.
We introduce \textbf{DASH},
a fast differentiable search framework for hybrid attention architecture
design, which relaxes discrete layer-wise attention operator placement into continuous
architecture logits,
prepares reusable teacher-aligned linear candidates,
and performs architecture-only search with model and operator weights
frozen to significantly enhance search efficiency.
On Qwen2.5-3B-Instruct,
DASH consistently outperforms a comprehensive suite of existing
selector-style hybrid attention design baselines,
showing that direct differentiable search can discover stronger
hybrid architectures.
Moreover,
DASH achieves stronger RULER performance than released Jet-Nemotron models
while remaining competitive on overlapping short-context and
general benchmarks.
Notably, each DASH search run uses only 12.3M tokens and takes about 20 minutes
on a single RTX Pro 6000 GPU,
corresponding to merely 0.006\% of the PostNAS search tokens reported by
Jet-Nemotron.
These results suggest that high-quality hybrid attention architectures can
be obtained through minutes-level differentiable search, providing a promising direction for hybrid architecture design.
\end{abstract}

\begin{figure*}[t]
    \centering
    \includegraphics[
        width=\textwidth
    ]{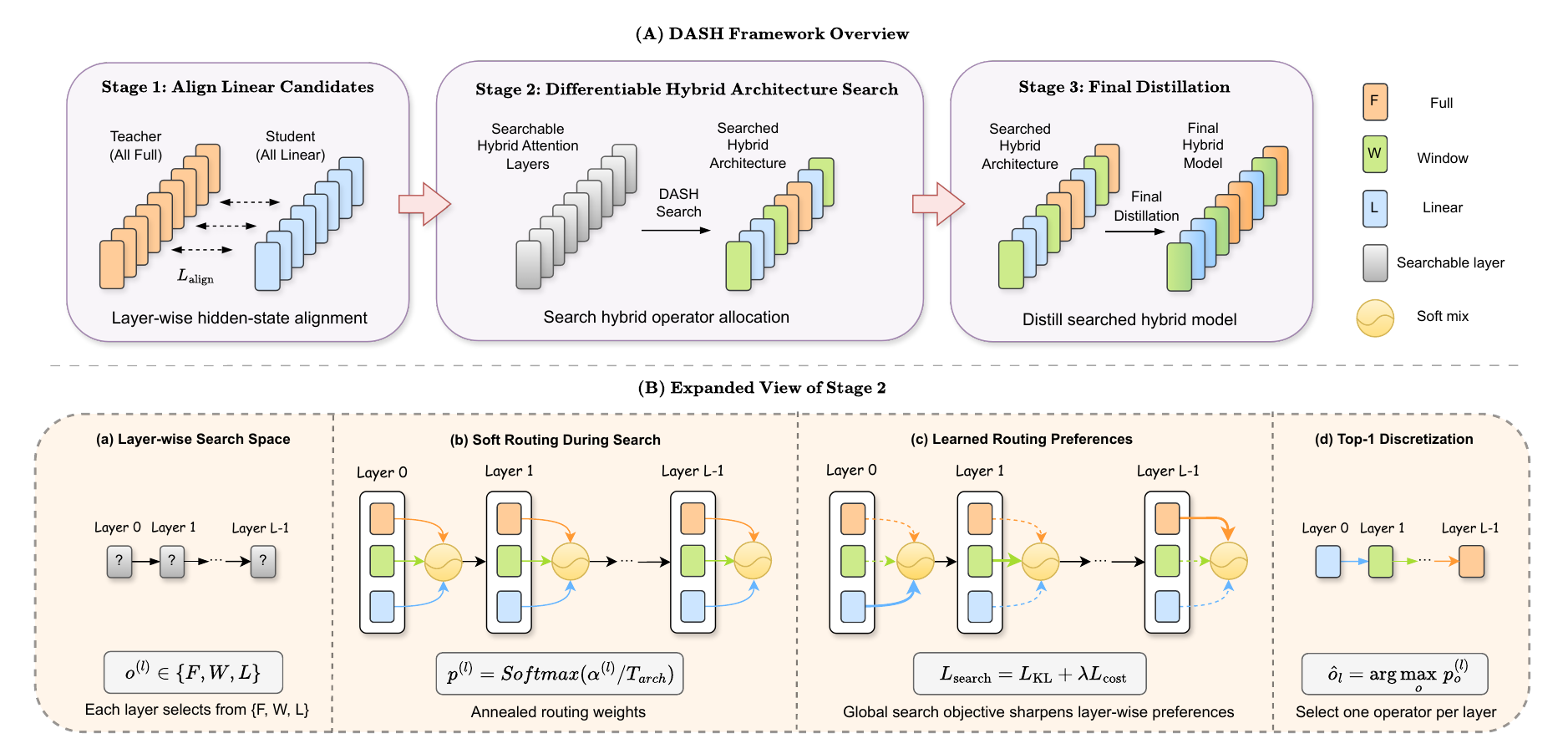}
    \caption{
        Overview of DASH.
        DASH turns hybrid attention design into an architecture-only
        differentiable search problem over layer-wise operator assignments.
        After Stage~1 prepares reusable teacher-aligned \textsc{Linear}
        candidates,
        Stage~2 relaxes each searchable layer into soft routing over
        \textsc{Full},
        \textsc{Window},
        and
        \textsc{Linear}
        operators and optimizes only architecture logits.
        Stage 3 then distills the selected discrete architecture into the final hybrid model used for evaluation and deployment.
    }
    \label{fig:dash_overview}
\end{figure*}

\section{Introduction}
\label{sec:introduction}

Transformer-based large language models (LLMs) 
\citep{Transformer}
rely heavily on full self-attention,
whose quadratic time and memory cost makes long-context inference
expensive.
In contrast,
efficient sequence mixers,
including linear attention 
and recurrent-state alternatives 
\citep{Transformers_are_RNNS,
mamba_linear_time_sequence_modeling_with_selective_state_spaces,
kimi_linear_an_expressive_efficient_attention_architecture},
scale more favorably,
but can sacrifice retrieval-oriented long-context behavior
\citep{repeat_after_me_transformers_are_better_than_state_space_models_at_copying}.
Hybrid architectures therefore offer a practical compromise:
some layers retain full attention,
while others use cheaper operators
\citep{nemotron3_efficient_and_open_intelligence,
jamba_hybrid_transformer_mamba_language_models,
qwen3_next}.
As hybrid architectures become an increasingly important paradigm for
efficient LLMs,
their architecture design becomes a central problem.

Early manual designs instantiate hybrid architectures with fixed
patterns,
such as interleaving expensive and cheap operators at prescribed
intervals
\citep{a_systematic_analysis_of_hybrid_linear_attention}.
While simple,
these rules ignore layer-dependent importance when deciding where
expensive operators should be placed.
Selector-style methods address this issue by estimating layer importance
or marginal utility from proxy signals
\citep{distill_then_replace_efficient_task_specific_hybrid_attention_model_construction, 
zebra_llama_towards_extremely_efficient_hybrid_models, 
distilling_to_hybrid_attention_models_via_kl_guided_layer_selection}.
However,
because operator choices interact across layers,
independently preferred layers may not compose into the best
complete architecture.

NAS-style search provides a more direct route by searching for the
architecture itself.
Jet-Nemotron validates this direction for hybrid LLM design
\citep{jet_nemotron_efficient_language_model_with_post_neural_architecture_search},
but its 200B-token PostNAS search cost limits its use as a routine
design method.
This raises the central question of 
whether NAS-style hybrid architecture search 
can be made cheap enough to serve as a routine design tool.

To address this gap,
we present \textbf{DASH},
a fast differentiable search framework for hybrid attention architecture
design.
DASH formulates architecture discovery as differentiable layer-wise
operator allocation and separates architecture search from final model
adaptation.
As a result,
DASH turns NAS-style hybrid architecture search from a costly specialized pipeline 
into a single-GPU procedure that completes within minutes.

Our contributions can be summarized as follows:
\begin{itemize}
    \item
    We introduce DASH,
    a differentiable search framework for hybrid attention architecture
    design.
    DASH moves beyond manual rules and selector-style architecture
    construction by directly searching layer-wise operator allocations
    in a differentiable NAS-style formulation.

    \item
    We make NAS-style hybrid architecture search practical through
    architecture-only differentiable optimization.
    DASH optimizes only architecture parameters with model and operator
    weights frozen:
    each search run uses only 12.3M tokens and completes in about
    20 minutes on a single RTX Pro 6000 GPU.

    \item
    We show that DASH finds strong hybrid architectures under comparable
    architecture budgets.
    On RULER,
    DASH consistently outperforms a comprehensive suite of existing
    selector-style hybrid attention design baselines,
    while requiring only minutes-level search.
    The searched models also achieve stronger RULER performance than
    released Jet-Nemotron models while remaining competitive on overlapping
    general benchmarks.
\end{itemize}

\section{Related Work}
\label{sec:related_work}

\noindent \textbf{Transformer-to-Hybrid Conversion}\hspace{0.5em}
A growing line of work studies how to transform pretrained
full-attention Transformers into more efficient architectures
without retraining language models from scratch
\citep{lolcats_on_low_rank_linearizing_of_large_language_models, 
radlads_rapid_attention_distillation_to_linear_attention_decoders_at_scale, 
finetuning_pretrained_transformers_into_rnns}.
Representative methods use weight transfer,
hidden-state alignment,
distillation,
and continual adaptation
to obtain efficient or hybrid students from full-attention teachers
\citep{the_mamba_in_the_llama_distilling_and_accelerating_hybrid_models, 
zebra_llama_towards_extremely_efficient_hybrid_models}.
These works establish conversion as a practical setting for efficient LLM construction, 
but leave open how the hybrid architecture itself should be designed.

\noindent \textbf{Hybrid Architecture Design and Layer Selection}\hspace{0.5em}
Layer-wise operator allocation is central to hybrid architecture design.
Selector-style methods estimate layer importance
or replacement sensitivity using task-guided probes
\citep{test_time_scaling_meets_associative_memory_challenges_in_subquadratic_models},
model-signal proxies, 
or teacher--student objectives
\citep{shortened_llama_depth_pruning_for_large_language_models_with_comparison_of_retraining_methods, 
zebra_llama_towards_extremely_efficient_hybrid_models,
distill_then_replace_efficient_task_specific_hybrid_attention_model_construction}.
Among these methods,
GA-S2 proposes a KL-guided teacher--student selection criterion for binary
\textsc{Full}/\textsc{Linear} allocation
and reports results across a broad suite of layer-selection strategies
\citep{distilling_to_hybrid_attention_models_via_kl_guided_layer_selection}.
While effective,
these methods still build the final architecture from layer-wise or
marginal signals.
Consequently,
they do not directly optimize the coupled operator assignment that
defines the final hybrid architecture and strongly affects model quality.

\noindent \textbf{Neural Architecture Search}\hspace{0.5em}
Neural architecture search (NAS) automates architecture design 
by searching over candidate operations,
connections,
or architectural configurations
\citep{neural_architecture_search_with_reinforcement_learning,
regularized_evolution_for_image_classifier_architecture_search}.
For hybrid LLM design,
Jet-Nemotron shows that NAS-style exploration can improve hybrid
architectures through a broad multi-stage PostNAS pipeline
\citep{jet_nemotron_efficient_language_model_with_post_neural_architecture_search}.
This effectiveness comes with substantial search cost:
its PostNAS search stages use 200B tokens,
reflecting the cost of exploring multiple discrete design stages and
evaluating or adapting candidate architectures with large token budgets.
Differentiable NAS offers a lighter alternative by relaxing discrete
architecture choices into continuous parameters that can be optimized
by gradient descent
\citep{darts_differentiable_architecture_search,
fair_darts_eliminating_unfair_advantages_in_differentiable_architecture_search,
zeroless_darts_improved_differentiable_architecture_search_with_refined_search_operation_and_early_stopping}.
DASH adopts this differentiable-search principle for hybrid attention
architecture design,
but restricts the search to layer-wise operator allocation and updates
only architecture logits,
with model and operator weights frozen during search.

\section{Preliminaries}
\label{sec:preliminaries}

\noindent \textbf{Notation and Transformer Blocks}\hspace{0.5em}
Let
$X^{(l)} \in \mathbb{R}^{T \times d}$
denote the hidden states
at layer $l$,
where $T$ is the sequence length
and $d$ is the model width.
We consider a decoder-only Transformer
with $L$ layers,
indexed by
$l \in \{0, 1, \dots, L-1\}$.
Each layer consists of
a sequence-mixing module
followed by a feed-forward network.
For a generic mixer
$\mathrm{Mix}^{(l)}(\cdot)$,
the layer update can be written as
\begin{equation}
U^{(l)}
=
X^{(l)}
+
\mathrm{Mix}^{(l)}
\bigl(
\mathrm{LN}(X^{(l)})
\bigr),
\qquad
X^{(l+1)}
=
U^{(l)}
+
\mathrm{FFN}^{(l)}
\bigl(
\mathrm{LN}(U^{(l)})
\bigr),
\label{eq:generic_block}
\end{equation}
where $\mathrm{LN}$ denotes layer normalization.

\noindent \textbf{Full and Window Attention}\hspace{0.5em}
Full attention
and window attention
share the same attention form,
differing only in the choice
of attention mask.
For attention-based operators,
queries, keys, and values
are computed as
\begin{equation}
Q = X W_Q,
\qquad
K = X W_K,
\qquad
V = X W_V,
\end{equation}
where
$W_Q, W_K, W_V \in \mathbb{R}^{d \times d_h}$
are projection matrices,
and $d_h$ is the head dimension
for a single attention head.
Ignoring multi-head notation for simplicity,
we write the generic attention operator as
\begin{equation}
\mathrm{Attn}(X; M)
=
\mathrm{Softmax}
\left(
\frac{QK^\top}{\sqrt{d_h}} + M
\right)
V,
\label{eq:attn_generic}
\end{equation}
where $M$ is an additive attention mask.
Standard causal full attention
uses the usual causal mask,
yielding
\begin{equation}
\mathrm{FullAttn}(X)
=
\mathrm{Attn}(X; M_{\mathrm{causal}}).
\label{eq:full_attn}
\end{equation}
Window attention instead restricts
each token to a local causal window
\citep{longformer_the_long_document_transformer, sliding_window_attention_training_for_efficient_large_language_models}.
Given a window size $w$,
token $t$ is allowed to attend only to tokens
in the range
$\max(1, t-w+1), \dots, t$.
This is implemented
by replacing the standard causal mask
with a local causal mask
$M_{\mathrm{win}}$,
yielding
\begin{equation}
\mathrm{WindowAttn}(X; w)
=
\mathrm{Attn}(X; M_{\mathrm{win}}).
\label{eq:window_attn}
\end{equation}
Full attention provides global token--token interaction
with quadratic cost,
whereas window attention preserves explicit local interaction
within a bounded receptive field.

\noindent \textbf{Linear Attention}~
Linear attention refers to
efficient sequence mixers
designed to scale linearly
with context length.
For notational simplicity,
we describe a linear attention layer
in recurrent form as
\begin{equation}
q_t = x_t W_Q,
\qquad
k_t = x_t W_K,
\qquad
v_t = x_t W_V,
\end{equation}
\begin{equation}
S_t = f_t(S_{t-1}, k_t, v_t),
\qquad
y_t = g_t(q_t, S_t),
\label{eq:linear_attn}
\end{equation}
where $S_t$ denotes the recurrent state,
and $f_t(\cdot)$ and $g_t(\cdot)$
depend on the specific
linear-attention instantiation.
The key distinction from
full or window attention
is that the history is represented
through a compressed state $S_t$,
rather than explicit token--token interaction.

\noindent \textbf{Hybrid Models as Layer-wise Operator Assignments}\hspace{0.5em}
A hybrid model is defined
by assigning a sequence operator
to each layer.
Let
$o_l \in
\{\textsc{Full}, \textsc{Window}, \textsc{Linear}\}$
denote the operator type
used at layer $l$.
Then the sequence operator at layer $l$ is
\begin{equation}
\mathrm{Operator}^{(l)}
=
\begin{cases}
\mathrm{FullAttn}^{(l)},
& o_l = \textsc{Full}, \\[2pt]
\mathrm{WindowAttn}^{(l)},
& o_l = \textsc{Window}, \\[2pt]
\mathrm{LinearAttn}^{(l)},
& o_l = \textsc{Linear}.
\end{cases}
\label{eq:hybrid_operator}
\end{equation}
Equivalently,
a hybrid architecture
can be represented
as a length-$L$ operator sequence
\begin{equation}
o
=
(o_0, o_1, \dots, o_{L-1}),
\qquad
o_l \in
\{\textsc{Full}, \textsc{Window}, \textsc{Linear}\}.
\label{eq:hybrid_arch}
\end{equation}
This representation makes explicit
that the hybrid architecture design problem
in this paper
is a layer-wise operator allocation problem.

\section{Method}
\label{sec:method}

\subsection{Overview}
\label{sec:overview}

Following the hybrid architecture formulation
in Eq.~\eqref{eq:hybrid_arch},
our goal is to learn a layer-wise operator assignment
$o = (o_0, o_1, \dots, o_{L-1})$
for an efficient hybrid student derived from a 
full-attention teacher.
In the tri-state setting,
each layer chooses from
$\mathcal{O}=\{\textsc{Full},\textsc{Window},\textsc{Linear}\}$,
and the binary
\textsc{Full}/\textsc{Linear}
setting is recovered by using
$\mathcal{O}=\{\textsc{Full},\textsc{Linear}\}$.

DASH addresses this allocation problem with a differentiable
architecture-search framework, implemented through the three stages
illustrated in Figure~\ref{fig:dash_overview}.
Stage~1 prepares teacher-aligned \textsc{Linear} candidates for each
layer.
Stage~2 is the central search stage:
it optimizes differentiable architecture parameters over the candidate
operators while keeping model and operator weights fixed.
Stage~3 instantiates the searched discrete architecture and distills it
into the final deployable hybrid model.

\subsection{Stage 1: Hidden-State Alignment of Linear Modules}
\label{sec:stage_1_hidden_state_alignment_of_linear_modules}

We first construct an all-linear student
by replacing each full-attention layer
in the teacher
with a linear-attention counterpart.
Let
$U_{\mathrm{T}}^{(l)}$
and
$U_{\mathrm{linear}}^{(l)}$
denote the post-mixer hidden states
of the teacher
and the all-linear student
at layer $l$, respectively.
We align the converted linear layer
to its teacher counterpart
with the layer-wise objective
\begin{equation}
\mathcal{L}_{\mathrm{align}}
=
\sum_{l=0}^{L-1}
\frac{1}{T}
\left\|
U_{\mathrm{T}}^{(l)}
-
U_{\mathrm{linear}}^{(l)}
\right\|_F^2,
\label{eq:align_loss}
\end{equation}
where $T$ is the sequence length.
During this stage,
only the linear modules are updated.
This stage produces teacher-aligned
linear modules
that serve as reusable layer candidates
for both architecture search and final hybrid model construction.

\subsection{Stage 2: Differentiable Hybrid Architecture Search}
\label{sec:stage_2_differentiable_hybrid_architecture_search}

\subsubsection{Continuous Layer-wise Relaxation}
\label{sec:continuous_layer_wise_relaxation}

We now search for a hybrid architecture
over the candidate operator set
$\mathcal{O}$.
At each layer,
the \textsc{Full} candidate is the original
full-attention layer from the teacher.
The \textsc{Window} candidate,
when included,
reuses the same attention weights
with a local causal mask.
The \textsc{Linear} candidate is the aligned
linear module obtained in Stage~1.

For each layer $l$,
we assign a learnable architecture logit vector
$\boldsymbol{\alpha}^{(l)} \in \mathbb{R}^{|\mathcal{O}|}$.
The corresponding routing probabilities are
\begin{equation}
\mathbf{p}^{(l)}
=
\mathrm{Softmax}
\bigl(
\boldsymbol{\alpha}^{(l)}
\bigr),
\qquad
p_o^{(l)}
\in
\mathbf{p}^{(l)},
\quad
o \in \mathcal{O}.
\label{eq:routing_probs}
\end{equation}
During search,
the mixer output at layer $l$
is defined as a soft combination
of candidate operators:
\begin{equation}
\mathrm{Mix}_{\mathrm{soft}}^{(l)}(X)
=
\sum_{o \in \mathcal{O}}
p_o^{(l)}
\,
\mathrm{Operator}_{o}^{(l)}(X),
\label{eq:soft_mixer}
\end{equation}
where
$\mathrm{Operator}_{o}^{(l)}$
denotes the layer-$l$ candidate operator
of type $o$.

Stage~2 is an architecture-only search stage:
all model and operator weights are frozen,
and only the architecture logits
$\boldsymbol{\alpha}$
are updated.
This avoids search-time weight adaptation
and substantially reduces the cost of architecture search.

\subsubsection{Teacher-Preserving and Cost-Regularized Search Loss}
\label{sec:teacher_preserving_and_cost_regularized_search_loss}

We optimize the architecture logits with a teacher-preserving objective
and a cost-regularization term that biases the search toward lower-cost
architectures.

To preserve teacher behavior,
we match the student output distribution
to that of the teacher.
Let
$\mathbf{z}_{\mathrm{T},t}$
and
$\mathbf{z}_{\mathrm{S},t}$
denote the teacher
and student logits
at token position $t$.
We define
\begin{equation}
\mathcal{L}_{\mathrm{KL}}
=
\frac{\tau^2}{T}
\sum_{t=1}^{T}
\mathrm{KL}
\left(
\mathrm{Softmax}
\left(
\frac{\mathbf{z}_{\mathrm{T},t}}{\tau}
\right)
\;\middle\|\;
\mathrm{Softmax}
\left(
\frac{\mathbf{z}_{\mathrm{S},t}}{\tau}
\right)
\right),
\label{eq:kl_loss}
\end{equation}
where $\tau$
is the distillation temperature.

Optimizing only
$\mathcal{L}_{\mathrm{KL}}$
would favor full attention,
since it is closest to
the teacher architecture.
We therefore add a normalized
relative attention-cost regularizer.
For the tri-state search space,
we use
\begin{equation}
\mathbf{c}
=
\bigl(
c_{\textsc{Full}},
c_{\textsc{Window}},
c_{\textsc{Linear}}
\bigr)
=
\left(
1,\ \frac{w}{T},\ 0
\right),
\label{eq:cost_vector_method}
\end{equation}
where $T$ is the sequence length
used during search
and $w$ is the window size.
This proxy measures cost relative
to full attention:
window attention contributes
in proportion to its local window ratio,
and linear attention is assigned zero cost as the reference operator
in this relative proxy.

Under the soft routing probabilities,
the expected architecture cost is
\begin{equation}
\mathcal{L}_{\mathrm{cost}}
=
\sum_{l=0}^{L-1}
\mathbf{p}^{(l)}
\cdot
\mathbf{c}.
\label{eq:cost_loss}
\end{equation}
The overall search loss is
\begin{equation}
\mathcal{L}_{\mathrm{search}}
=
\mathcal{L}_{\mathrm{KL}}
+
\lambda
\mathcal{L}_{\mathrm{cost}},
\label{eq:search_loss}
\end{equation}
where $\lambda$
is the cost regularization coefficient.
The first term encourages teacher preservation,
while the second term biases the search
toward cheaper architectures.

\subsubsection{Annealed Relaxation}
\label{sec:annealed_relaxation}

The relaxation above makes
layer-wise operator allocation differentiable.
However,
the final model is discrete,
so diffuse routing distributions can make
top-1 discretization unstable.
To reduce this soft-to-hard mismatch,
we use a temperature-scaled variant of
Eq.~\eqref{eq:routing_probs}:
\begin{equation}
\mathbf{p}^{(l)}
=
\mathrm{Softmax}
\left(
\frac{
\boldsymbol{\alpha}^{(l)}
}{
T_{\mathrm{arch}}
}
\right),
\label{eq:annealed_routing_probs}
\end{equation}
where $T_{\mathrm{arch}}$
is annealed 
from a higher value to a lower value
during search.
The routing distribution is smoother
early in search
and becomes sharper
near discretization.

\subsubsection{Discretization}
\label{sec:discretization}

After search,
we derive a discrete architecture
by selecting the dominant operator
at each layer:
\begin{equation}
\hat{o}_l
=
\arg\max_{o \in \mathcal{O}}
p_o^{(l)}.
\label{eq:argmax_discretization}
\end{equation}
The selected operators are then used
to instantiate the hybrid student
for the final distillation stage.

\subsection{Stage 3: Final Distillation}
\label{sec:stage_3_final_distillation}

Given the searched discrete architecture,
we instantiate the corresponding hybrid student
and distill it from the teacher
to obtain the final deployable model.
At this stage,
the search-time soft mixtures are removed,
and each layer executes exactly one operator.
All student parameters are trained
with the same output-level teacher KL objective,
allowing the model to adapt
to the searched architecture.

\section{Experiments}
\label{sec:experiments}

\subsection{Experimental Setup}
\label{sec:experimental_setup}

\noindent \textbf{Model and Operators} \hspace{0.5em}
We convert
\texttt{Qwen2.5-3B-Instruct}
\citep{qwen2_5_technical_report},
a 36-layer full-attention Transformer,
into hybrid students.
\textsc{Linear} is instantiated as Gated DeltaNet
\citep{gated_delta_networks_improving_mamba2_with_delta_rule},
and \textsc{Window} uses window size $w=512$.

\noindent \textbf{Implementation Details} \hspace{0.5em}
Appendix~\ref{app:training_search_config} provides the training and search configurations for the main experimental setup. 
Appendix~\ref{app:layer0_handling} describes the fixed layer-0 convention used by DASH:
layer 0 is held as \textsc{Full} during Stage~2 search and instantiated
as \textsc{Linear} in the final Stage~3 model.

\noindent \textbf{Training and Search Configuration} \hspace{0.5em}
All stages use the DCLM generic-text corpus
\citep{datacomp_lm_in_search_of_the_next_generation_of_training_sets_for_language_models}.
Stage~1 uses approximately 100M tokens
to prepare aligned \textsc{Linear} candidates,
and Stage~3 distills each final hybrid model
for approximately 600M tokens.
Each DASH Stage~2 search run uses 12.3M tokens.
Further training and search details are provided in
Appendix~\ref{app:training_search_config}.

\noindent \textbf{Benchmarks} \hspace{0.5em}
We evaluate retrieval-oriented long-sequence capability
with RULER
\citep{ruler_whats_the_real_context_size_of_your_long_context_language_models}
at 4096 context length.
We also evaluate short-context capability with MMLU
\citep{measuring_massive_multitask_language_understanding},
ARC-Easy and ARC-Challenge
\citep{think_you_have_solved_question_answering_try_arc_the_ai2_reasoning_challenge},
PIQA
\citep{piqa_reasoning_about_physical_commonsense_in_natural_language},
and WinoGrande
\citep{winogrande_an_adversarial_winograd_schema_challenge_at_scale}.

\noindent \textbf{Attention Budget Definition} \hspace{0.5em}
For binary models,
the budget is the number of full-attention layers.
For tri-state models,
we use sequence length $T=4096$ and window size $w=512$,
so the Stage~2 relative cost vector is
$c=(1.0,0.125,0.0)$
for
$(\textsc{Full},\textsc{Window},\textsc{Linear})$,
and the final realized budget after hard discretization is
$B=n_{\textsc{Full}}+0.125\,n_{\textsc{Window}}$.

\subsection{Main Results}
\label{sec:main_results}

\begin{table*}[!t]
    \centering
    \setlength{\tabcolsep}{5pt}
    \renewcommand{\arraystretch}{0.96}
    \caption{
        RULER comparison on Qwen2.5-3B-Instruct against the broad
        \textsc{Full}/\textsc{Linear}
        selector landscape reported by GA-S2
        \citep{distilling_to_hybrid_attention_models_via_kl_guided_layer_selection}.
        DASH rows are from our 600M-token Stage~3 runs,
        matching the final-distillation token count reported for GA-S2.
        For DASH-3,
        the realized budgets are
        4.875,
        9.0,
        and
        17.75,
        respectively.
        Boldface denotes the best result,
        and underlining denotes the second-best result for each reported budget
        setting.
    }
    \label{tab:ruler_selector_landscape}
    \begin{tabular}{llccc}
        \toprule
        Selector
        & Type
        & $Budget=5$
        & $Budget=9$
        & $Budget=18$ \\
        \midrule
        Uniform
        & Fixed
        & 44.12
        & 69.04
        & 80.31 \\
        KV
        & Task probe
        & 25.43
        & 75.39
        & 82.57 \\
        AR
        & Task probe
        & 54.17
        & 63.85
        & 87.17 \\
        VT
        & Task probe
        & 29.22
        & 47.80
        & 74.09 \\
        CWE
        & Task probe
        & 29.00
        & 49.07
        & 84.44 \\ 
        AR-MH
        & Task probe
        & 43.71
        & 71.60
        & 79.83 \\
        Act-MSE
        & Model proxy
        & 38.27
        & 44.21
        & 71.75 \\
        LM-PPL
        & Model proxy
        & 41.29
        & 51.00
        & 69.97 \\
        SMART
        & Prior
        & 40.89
        & 64.01
        & 81.90 \\
        GR-S1
        & GR (S1-MSE)
        & 35.63
        & 48.27
        & 82.09 \\
        GA-S1
        & GA (S1-MSE)
        & 38.43
        & 54.08
        & 78.73 \\
        Avg-S1
        & Avg. (S1-MSE)
        & 39.60
        & 49.33
        & 82.26 \\
        GR-S2
        & GR (S2-KL)
        & 58.04
        & 82.59
        & 88.69 \\
        GA-S2
        & GA (S2-KL)
        & 66.17
        & 86.31
        & 90.61 \\
        Avg-S2
        & Avg. (S2-KL)
        & 70.75
        & 82.05
        & 90.51 \\
        \midrule
        \rowcolor{dashhighlight}
        DASH-2
        & Diff. search
        & \underline{72.99}
        & \underline{86.88}
        & \underline{91.14} \\
        \rowcolor{dashhighlight}
        DASH-3
        & Diff. search
        & \textbf{73.58}
        & \textbf{87.63}
        & \textbf{91.42} \\
        \bottomrule
    \end{tabular}
\end{table*}

\noindent \textbf{RULER Selector Landscape}\hspace{0.5em}
Table~\ref{tab:ruler_selector_landscape}
compares DASH with a comprehensive suite of existing selector-style
hybrid attention design baselines from the GA-S2 reported
\textsc{Full}/\textsc{Linear}
selector landscape on Qwen2.5-3B-Instruct.
The baseline rows are reported by
\citet{distilling_to_hybrid_attention_models_via_kl_guided_layer_selection},
whereas DASH rows are from our experiments.

DASH outperforms all reported selector baselines at all three budget
levels.
In the same binary
\textsc{Full}/\textsc{Linear}
space,
DASH-2 already exceeds the best reported selector in every budget
column.
Expanding the candidate set to
\textsc{Full}/\textsc{Window}/\textsc{Linear},
DASH-3 further improves the quality--budget trade-off.
These results show that direct differentiable architecture search can
find stronger hybrid architectures than selector-style construction from
layer-wise signals,
and that DASH can naturally benefit from a heterogeneous operator space.

\begin{table*}[!t]
    \centering
    \setlength{\belowcaptionskip}{6pt}
    \setlength{\tabcolsep}{4pt}
    \renewcommand{\arraystretch}{0.98}
    \small
    \caption{
External reference comparison with released Jet-Nemotron models.
Scores are measured by us under the same evaluation setup.
DASH rows are Qwen2.5-3B-Instruct models using DASH-3 architectures
and 600M-token Stage~3 distillation.
Search and Train report token counts:
for DASH,
Search is per Stage~2 run and Train is 100M Stage~1 plus 600M Stage~3;
for Jet-Nemotron,
costs follow reported PostNAS token counts.
N/R marks an unreported token count.
    }
    \label{tab:jet_nemotron_reference}
    \begin{tabular}{lcccccccc}
    \toprule
    Model
    & Search 
    & Train 
    & RULER
    & MMLU
    & ARC-E
    & ARC-C
    & PIQA
    & WinoGrande \\
    \midrule
    DASH-3 ($B=4.875$)
    & 12.3M / run
    & 700M
    & 73.58
    & 57.41
    & 76.68
    & 51.88
    & 77.75
    & 67.09 \\
    DASH-3 ($B=9.0$)
    & 12.3M / run
    & 700M
    & 87.63
    & 62.55
    & 76.85
    & 51.79
    & 77.86
    & 69.53 \\
    DASH-3 ($B=17.75$)
    & 12.3M / run
    & 700M
    & 91.42
    & 63.95
    & 76.98
    & 51.62
    & 78.07
    & 68.98 \\
    \midrule
    Jet-Nemotron-2B
    & 200B
    & 400B
    & 83.86
    & 60.87
    & 74.79
    & 48.89
    & 75.14
    & 65.51 \\
    Jet-Nemotron-4B
    & N/R
    & 400B
    & 83.71
    & 65.24
    & 79.25
    & 51.28
    & 77.97
    & 70.17 \\
    \bottomrule
    \end{tabular}
\end{table*}

\noindent \textbf{External NAS-Style Hybrid Model Reference} \hspace{0.5em}
Table~\ref{tab:jet_nemotron_reference}
places DASH-3 in the broader context of NAS-style hybrid LLM design by
comparing it with released Jet-Nemotron models.
All results are measured by us under the same
evaluation setup,
but the comparison is not controlled because the models differ in
backbone,
scale,
architecture,
search procedure,
and training pipeline.

DASH-3 achieves stronger RULER performance than both Jet-Nemotron models
at the medium and high budget regimes,
while remaining competitive on overlapping short-context and general
benchmarks.
The cost contrast is also substantial:
DASH uses 12.3M tokens per search run and 700M training tokens per final
model,
whereas Jet-Nemotron reports 200B PostNAS search tokens and 400B
training tokens for released models,
with the 4B search cost not separately reported.

Together,
Tables~\ref{tab:ruler_selector_landscape}
and~\ref{tab:jet_nemotron_reference}
show that DASH improves over selector-style layer allocation on the same
Qwen2.5 setting,
while producing hybrid models that remain competitive with heavyweight
NAS-style references at much smaller reported search and training cost.

\begin{figure*}[t]
    \centering
    \includegraphics[width=0.95\textwidth]{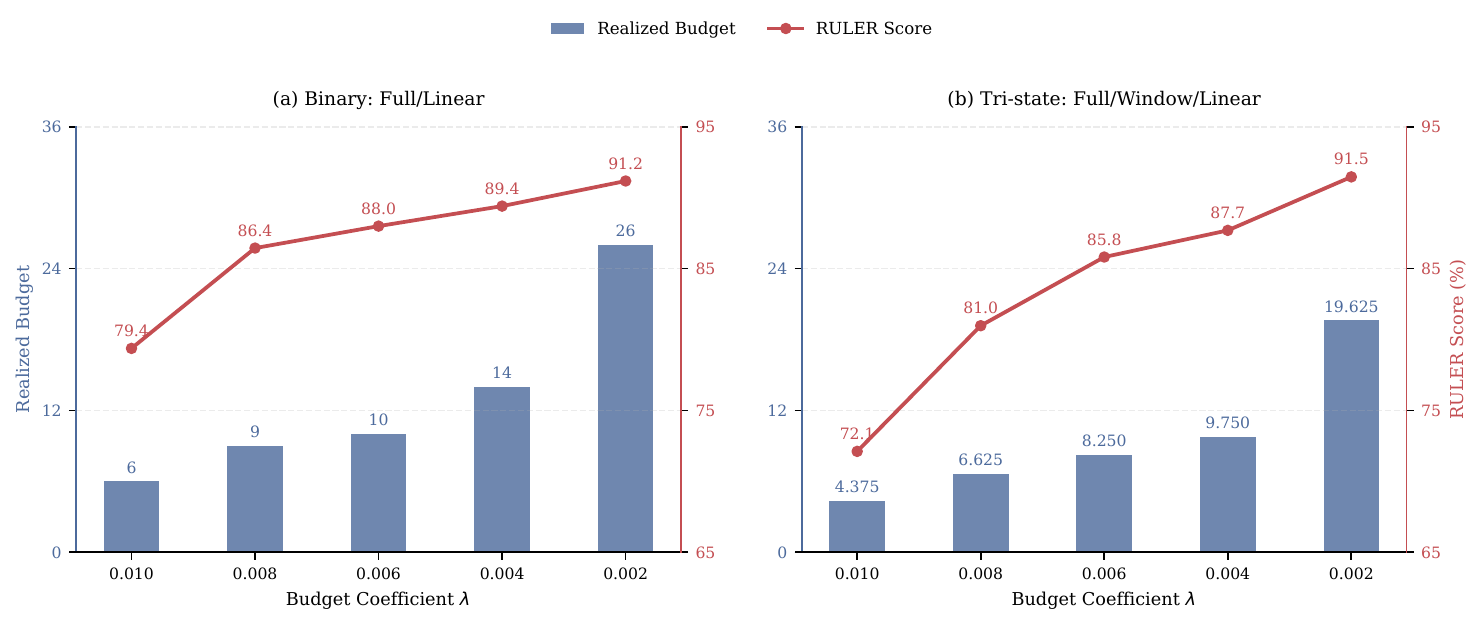}
    \caption{
        Effect of the budget coefficient $\lambda$.
        Bars show the realized budget,
        and lines show the final RULER score.
        Lower $\lambda$ weakens the cost penalty,
        leading to larger realized budgets
        and higher RULER scores
        in both the binary
        \textsc{Full}/\textsc{Linear}
        and tri-state
        \textsc{Full}/\textsc{Window}/\textsc{Linear}
        search spaces.
        Final RULER scores in this diagnostic sweep are measured after
        300M-token Stage~3 distillation.
    }
    \label{fig:lambda_ablation}
\end{figure*}

\subsection{Budget Control with \texorpdfstring{$\lambda$}{lambda}}
\label{sec:effect_of_lambda_on_budget_and_performance}

We vary the budget coefficient $\lambda$ in both the binary
\textsc{Full}/\textsc{Linear}
and tri-state
\textsc{Full}/\textsc{Window}/\textsc{Linear}
search spaces.
As shown in Figure~\ref{fig:lambda_ablation},
$\lambda$ acts as a practical control knob for tracing
quality--efficiency operating points:
larger $\lambda$ penalizes expensive operators more strongly
and yields lower-budget architectures,
while smaller $\lambda$ weakens the penalty and generally leads to
higher realized budgets and higher RULER scores.
Because each Stage~2 run is inexpensive,
sweeping $\lambda$ is practical for selecting operating points.
At the same time,
the mapping from $\lambda$ to realized budget is discrete after top-1
discretization,
so we report each model by its realized budget rather than 
using $\lambda$ itself as the operating point.

\begin{table}[t]
    \centering
    \small
    \setlength{\tabcolsep}{4pt}
    \caption{
Routing diagnostics for tri-state searches with and without annealing.
Paired rows use the same budget coefficient $\lambda$,
but their budgets are realized after top-1 discretization and are
therefore not exactly matched.
Routing metrics are computed from the final soft probabilities before
discretization.
RULER scores are auxiliary outcomes measured after 300M-token Stage~3
distillation.
    }
    \label{tab:anneal_ablation}
    \vspace{0.5em}
    \begin{tabular}{lccccccc}
        \toprule
        Regime
        & Setting
        & Budget
        & Avg. Ent. $\downarrow$
        & Avg. Top-1 $\uparrow$
        & Avg. Margin $\uparrow$
        & Ambig. $\downarrow$
        & RULER \\
        \midrule
        $\sim$5
        & w/o anneal
        & 4.250
        & 0.520
        & 0.753
        & 0.529
        & 9 
        & 68.90 \\
        $\sim$5
        & w/ anneal
        & 4.875
        & \textbf{0.323}
        & \textbf{0.832}
        & \textbf{0.671}
        & \textbf{4}
        & 73.21 \\
        \midrule
        $\sim$9
        & w/o anneal
        & 8.375
        & 0.628
        & 0.721
        & 0.497
        & 7 
        & 85.56 \\
        $\sim$9
        & w/ anneal
        & 9.000
        & \textbf{0.451}
        & \textbf{0.788}
        & \textbf{0.608}
        & \textbf{6} 
        & 87.47 \\
        \midrule
        $\sim$18
        & w/o anneal
        & 18.500
        & 0.745
        & 0.669
        & 0.438
        & 7 
        & 90.97 \\
        $\sim$18
        & w/ anneal
        & 17.750
        & \textbf{0.573}
        & \textbf{0.758}
        & \textbf{0.581}
        & \textbf{5} 
        & 91.30 \\
        \bottomrule
    \end{tabular}
\end{table}

\subsection{Annealing as a Discretization Stabilizer}
\label{sec:annealing_as_a_discretization_stabilizer}

We examine annealing as a discretization stabilizer
by comparing tri-state searches with and without annealing
under the same budget coefficient $\lambda$
at representative low,
medium,
and high budget regimes.
Table~\ref{tab:anneal_ablation}
reports diagnostics computed from the final soft routing probabilities
before hard discretization:
average entropy,
average top-1 probability,
average top-1 margin,
and the number of ambiguous layers whose top-1 margin is below 0.2.
Lower entropy indicates a more concentrated routing distribution;
higher top-1 probability and margin indicate a more decisive operator
choice;
and fewer ambiguous layers indicate fewer near-ties before
discretization.
Together,
these diagnostics measure the confidence of the soft-to-hard conversion.

Across all three regimes,
annealing lowers entropy,
increases top-1 probability and margin,
and reduces the number of ambiguous layers.
These trends indicate sharper routing distributions
and higher confidence before hard discretization.
We therefore use annealing to stabilize the conversion from the
differentiable soft architecture to the final discrete hybrid model.

\subsection{Inference Efficiency}
\label{sec:inference}

\begin{figure*}[!t]
    \centering
    \includegraphics[width=0.95\textwidth]{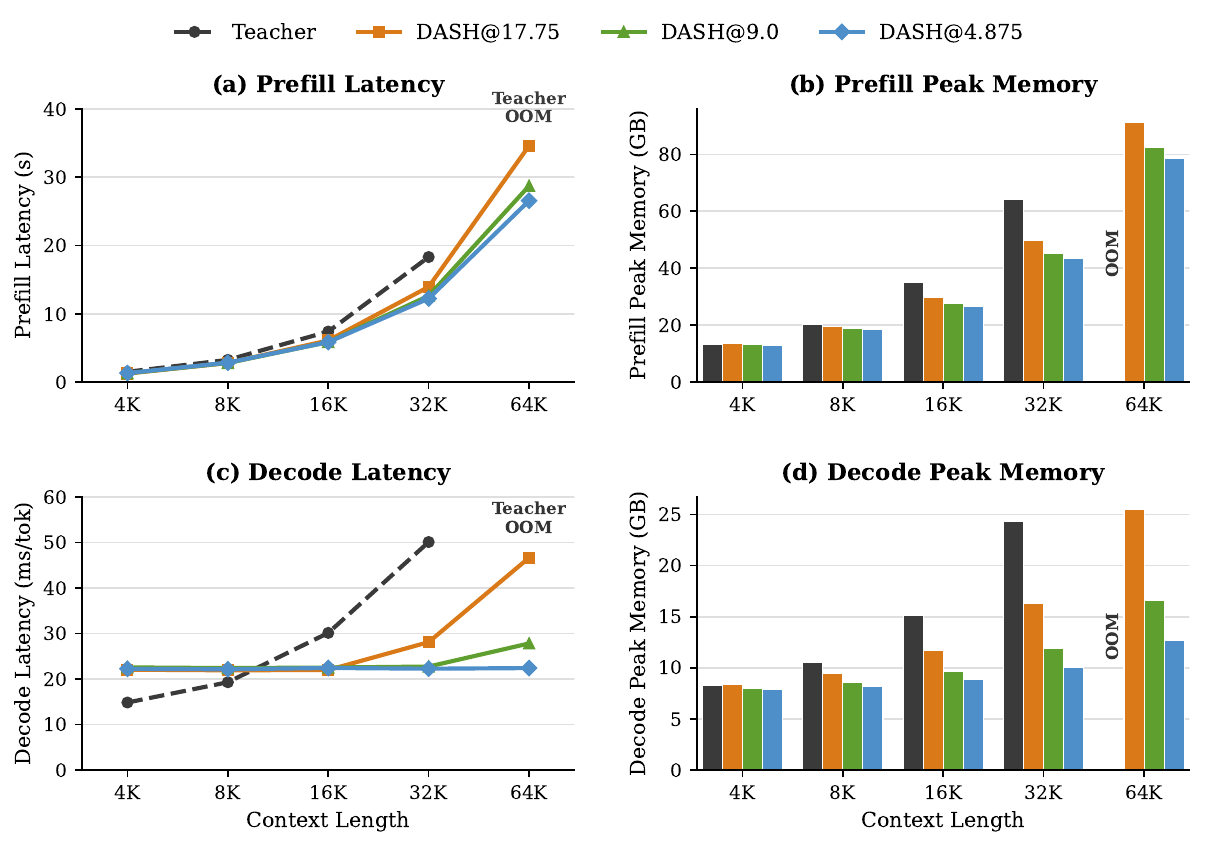}
    \caption{
        Inference efficiency at batch size 16.
        DASH@$B$ denotes a tri-state DASH model
        with realized budget $B$.
        The teacher runs out of memory at 64K context,
        while all DASH variants remain executable at that length.
    }
    \label{fig:inference_efficiency}
\end{figure*}

We measure prefill and decode latency,
as well as peak GPU memory,
at batch size 16
across long context lengths.
Figure~\ref{fig:inference_efficiency}
compares the full-attention teacher
with tri-state DASH models
at three
representative realized budgets under the same measurement setup.

At 32K context,
DASH@4.875 achieves
a 1.50$\times$ prefill speedup
and a 2.25$\times$ decode speedup
over the teacher,
while reducing prefill and decode peak memory
by 1.47$\times$ and 2.41$\times$, respectively.
At 64K context,
the teacher runs out of memory,
whereas all DASH variants remain executable.
As expected,
the efficiency gains are larger
for lower-budget DASH models
and become more pronounced
at longer context lengths,
where full attention incurs higher memory
and computation cost.

\section{Limitations and Future Work}
DASH focuses on low-cost differentiable search for layer-wise hybrid
attention allocation,
but our current validation is still limited by practical compute
constraints.
The main experiments use Qwen2.5-3B-Instruct as the backbone,
instantiate linear layers with Gated DeltaNet,
and search over a predefined operator set;
extending DASH to more model families,
larger scales,
and broader efficient-operator spaces remains important future work.
Similarly,
our final hybrid models are trained with 600M-token Stage~3 distillation,
which enables controlled architecture comparison but is not intended to
match the extensive post-search training budgets used by some released
hybrid LLMs.
This may partly explain why DASH preserves competitive general benchmark
performance but can still lag stronger or larger references on some
short-context tasks.
The current budget definition is a simple relative attention-cost proxy
based on operator type and window ratio;
future work could replace it with calibrated hardware-aware objectives
that directly model latency,
memory,
or device-specific constraints.
Finally,
the reported Stage~2 cost is measured per search run:
selecting a desired realized budget may require a small sweep over the
cost coefficient $\lambda$
because the final architecture is obtained by hard discretization.
Even with such sweeps,
the total search cost remains small relative to heavyweight NAS-style
pipelines.

\section{Conclusion}
\label{sec:conclusion}

We presented \textbf{DASH},
a fast differentiable search framework for hybrid attention architecture
design.
DASH shows that NAS-style hybrid architecture search need not rely on
heavyweight multi-stage pipelines:
by reducing architecture discovery to lightweight differentiable
operator allocation,
it turns NAS-style search from a costly specialized pipeline into a
routine experimental tool for hybrid LLM design.
Across Qwen2.5-3B-Instruct experiments,
DASH outperforms the reported GA-S2 selector landscape on RULER,
and the searched hybrid models achieve stronger retrieval-oriented
long-context performance than released Jet-Nemotron models while
remaining competitive on overlapping general benchmarks.
These results suggest a broader direction for efficient LLM design:
hybrid attention architectures should be searched directly,
not hand-designed or assembled from isolated layer scores,
and differentiable architecture search provides a practical path toward
that goal.

\bibliographystyle{unsrtnat}
\bibliography{references}


\FloatBarrier
\appendix

\section{Training and Search Details}
\label{app:training_search_config}

All training, search, and inference-measurement runs were conducted on
NVIDIA RTX PRO 6000 Blackwell GPUs.
For the main DASH results,
selecting the reported operating points additionally required lightweight
$\lambda$/seed sweeps;
across the main binary and tri-state searches,
we ran approximately 50 Stage~2 searches in total,
corresponding to less than 1B processed search tokens.
These additional sweeps are search-only runs and do not require training
a final Stage~3 model for every searched candidate.

Tables~\ref{tab:stage13_training_config}
and~\ref{tab:stage2_search_config}
summarize the Stage~1/3 training configurations
and the Stage~2 search configuration.
Table~\ref{tab:main_search_settings}
reports the search settings used for the main DASH operating points.
We use AdamW
\citep{decoupled_weight_decay_regularization}
for all optimization stages
and train in bfloat16 precision.

\begin{table}[!htbp]
\centering
\small
\caption{
Training configurations for Stage~1 and Stage~3.
}
\label{tab:stage13_training_config}
\begin{tabular}{lcc}
\toprule
 & Stage~1: Linear Alignment
 & Stage~3: Final Distillation \\
\midrule
Seq. Len.
& 512
& 4096 \\
Training tokens
& 100M
& 600M \\
GPUs
& 1
& 2 \\
Approx. wall-clock time
& $\sim$ 2.5 hours / once
& $\sim$ 10 hours / model \\
Global batch size
& 96
& 32 \\
Micro batch size
& 6 
& 2 \\
LR scheduler
& Cosine
& Constant \\
Non-attention LR
& --
& $7.0\times 10^{-6}$ \\
Attention-operator LR
& $1.0\times 10^{-3}$
& $1.0\times 10^{-3}$ \\
Weight decay
& 0.01
& 0.01 \\
Objective
& Hidden-state alignment
& Output-level KL distillation \\
\bottomrule
\end{tabular}
\end{table}

\begin{table}[!htbp]
\centering
\small
\caption{
Differentiable search configuration for Stage~2.
Only the architecture logits $\alpha$ are optimized during search.
Search tokens are counted over micro steps.
}
\label{tab:stage2_search_config}
\begin{tabular}{lc}
\toprule
Search parameter & Value \\
\midrule
Seq. Len.
& 4096 \\
Micro batch size
& 2 \\
Gradient accumulation steps
& 8 \\
Search micro steps
& 1500 \\
Processed search tokens
& 12.3M \\
GPUs
& 1 \\
Approx. wall-clock time
& $\sim$20 min / run \\
LR for architecture logits $\alpha$
& 0.1 \\
Weight decay on $\alpha$
& 0.0 \\
Initial $T_{\rm arch}$
& 1.0 \\
Final $T_{\rm arch}$
& 0.1 \\
$T_{\rm arch}$ annealing steps
& 1500 \\
\bottomrule
\end{tabular}
\end{table}

\begin{table}[!htbp]
\centering
\small
\caption{
Search configurations used for the main DASH operating points.
For binary DASH,
the realized budget equals the number of \textsc{Full} layers.
For tri-state DASH,
it is computed as
$n_{\rm FULL}+0.125n_{\rm WINDOW}$.
}
\label{tab:main_search_settings}
\begin{tabular}{lccc}
\toprule
Model
& Candidate space
& $\lambda$
& Realized budget \\
\midrule
DASH 2-candidate, low budget
& \{\textsc{Full}, \textsc{Linear}\}
& 0.012
& 5 \\
DASH 2-candidate, medium budget
& \{\textsc{Full}, \textsc{Linear}\}
& 0.008
& 9 \\
DASH 2-candidate, high budget
& \{\textsc{Full}, \textsc{Linear}\}
& 0.003
& 18 \\
DASH 3-candidate, low budget
& \{\textsc{Full}, \textsc{Window}, \textsc{Linear}\}
& 0.009
& 4.875 \\
DASH 3-candidate, medium budget
& \{\textsc{Full}, \textsc{Window}, \textsc{Linear}\}
& 0.0047
& 9.0 \\
DASH 3-candidate, high budget
& \{\textsc{Full}, \textsc{Window}, \textsc{Linear}\}
& 0.00234
& 17.75 \\
\bottomrule
\end{tabular}
\end{table}

\FloatBarrier
\section{Layer-0 Handling}
\label{app:layer0_handling}

In all DASH experiments,
we use a fixed convention for layer~0.
During Stage~2 search,
layer~0 is held as \textsc{Full}
and excluded from the searchable layers.
All other layers are searched normally
with learnable architecture logits.
This convention keeps layer 0 outside the optimized architecture
variables and is applied uniformly across all DASH runs,
without per-budget or per-result tuning.

When instantiating the final discrete architecture for Stage~3,
layer~0 is set to \textsc{Linear}.
Accordingly,
reported DASH budgets are computed from the final instantiated model,
where layer~0 contributes zero cost.

Routing diagnostics are computed over searchable layers only,
excluding layer~0.
Budget-sweep allocation visualizations follow the same convention,
since they analyze the learned searchable-layer assignments.
In contrast,
final discrete architecture summaries and visualizations include
layer~0 after applying the final instantiation convention.

\FloatBarrier
\section{Additional Evaluation Results}
\label{app:additional_evaluation_results}

\subsection{Selector Tags in the GA-S2 Reported Landscape}
\label{app:ga_s2_selector_tags}

We summarize the selector tags used in the GA-S2 reported RULER
landscape in Table~\ref{tab:ruler_selector_landscape}.
The reported results and detailed implementations follow
\citet{distilling_to_hybrid_attention_models_via_kl_guided_layer_selection}.
Here,
S1 and S2 follow the notation of GA-S2:
S1 denotes hidden-state/MSE-based signals,
whereas S2 denotes KL-guided distillation signals.
This notation is independent of the DASH stage numbering,
where Stage~2 refers to differentiable architecture search and
Stage~3 refers to final distillation.
We include this summary only to make the selector names self-contained.

\begin{table}[!htbp]
\centering
\footnotesize
\setlength{\tabcolsep}{3pt}
\captionsetup{skip=6pt}
\caption{
Summary of selector tags used in the GA-S2 reported RULER landscape.
S1 and S2 follow GA-S2 notation:
S1 denotes hidden-state/MSE-based signals,
and S2 denotes KL-guided distillation signals.
Detailed definitions and implementations follow GA-S2.
}
\label{tab:ga_s2_selector_tags}
\begin{tabular}{lll}
\toprule
Tag & Type & Short description \\
\midrule
Uniform & Fixed rule & Evenly places full-attention layers. \\
KV, AR, AR-MH, VT, CWE & Task-guided probes & Use synthetic recall or tracking tasks to score layers. \\
Act-MSE, LM-PPL & Model-signal proxies & Use activation mismatch or language-model perplexity. \\
SMART & Prior selector & Reported selector baseline included by GA-S2. \\
GR-S1 / GR-S2 & Greedy removal & Remove layers using GA-S2 S1-MSE or S2-KL signals. \\
GA-S1 / GA-S2 & Greedy addition & Add layers using GA-S2 S1-MSE or S2-KL signals. \\
Avg-S1 / Avg-S2 & Rank averaging & Average greedy rankings from removal and addition. \\
\bottomrule
\end{tabular}
\end{table}

\FloatBarrier
\subsection{Controlled 300M-token Training Comparison}
\label{app:controlled_300m_results}

Table~\ref{tab:controlled_300m_results}
provides a controlled comparison under our own training and evaluation
pipeline.
Unlike Table~\ref{tab:ruler_selector_landscape},
where the selector baselines are reported by GA-S2,
all rows in this table are trained and evaluated by us with
300M-token Stage~3 distillation.
For Uniform and GA-S2,
we use the layer allocations reported by GA-S2,
and then run the same Stage~1 preparation and Stage~3 distillation
pipeline used for DASH.

\begin{table}[!htbp]
    \centering
    \setlength{\tabcolsep}{5pt}
    \renewcommand{\arraystretch}{1.0}
    \small
    \caption{
        Controlled 300M-token training comparison on
        Qwen2.5-3B-Instruct.
        Uniform and GA-S2 instantiate the layer allocations reported by
        GA-S2,
        while DASH-2 and DASH-3 use architectures searched by DASH.
        All rows are trained and evaluated by us under the same
        300M-token Stage~3 distillation setup.
    }
    \label{tab:controlled_300m_results}
    \begin{tabular}{lcccccc}
        \toprule
        Regime
        & Binary budget
        & DASH-3 budget
        & Uniform
        & GA-S2
        & DASH-2
        & DASH-3 \\
        \midrule
        Low
        & 5
        & 4.875
        & 18.47
        & 58.08
        & \underline{72.51}
        & \textbf{73.21} \\
        Medium
        & 9
        & 9.0
        & 65.23
        & 85.50
        & \underline{86.44}
        & \textbf{87.47} \\
        High
        & 18
        & 17.75
        & 78.04
        & 90.58
        & \underline{90.79}
        & \textbf{91.30} \\
        \bottomrule
    \end{tabular}
\end{table}

DASH outperforms both reproduced Uniform and GA-S2 architectures across
all three budget regimes.
The binary DASH-2 results already exceed the reproduced GA-S2
architectures at matched
\textsc{Full}/\textsc{Linear}
budgets,
and DASH-3 further improves the results by using the expanded
\textsc{Full}/\textsc{Window}/\textsc{Linear}
operator space.

\FloatBarrier
\subsection{Full RULER Results}
\label{app:full_ruler_results}

Table~\ref{tab:ruler_full_results}
reports the full RULER breakdown at 4096 context length for models
evaluated by us.
The DASH rows correspond to the 600M-token Stage~3 models used in
Table~\ref{tab:ruler_selector_landscape}.
The Jet-Nemotron rows are evaluated from released checkpoints under the
same RULER protocol used in Table~\ref{tab:jet_nemotron_reference}.

\begin{table*}[!htbp]
\centering
\scriptsize
\setlength{\tabcolsep}{2.4pt}
\renewcommand{\arraystretch}{0.96}
\captionsetup{skip=6pt}
\caption{
Full RULER results at 4096 context length.
All rows are evaluated by us under the same RULER setup.
DASH rows use 600M-token Stage~3 distillation.
Jet-Nemotron rows are evaluated from released models,
and the Qwen2.5-3B-Instruct teacher is included as the full-attention
reference.
All values are percentages.
}
\label{tab:ruler_full_results}
\resizebox{\textwidth}{!}{
\begin{tabular}{lcccccccccccccc}
\toprule
Model
& Avg.
& MK1
& MK2
& MK3
& MQ
& MV
& S1
& S2
& S3
& CWE
& FWE
& HP
& SQ
& VT \\
\midrule
Qwen2.5-3B-Instruct
& 91.13
& 99.60
& 99.20
& 98.40
& 99.85
& 98.90
& 100.00
& 100.00
& 100.00
& 80.74
& 85.33
& 50.00
& 72.75
& 99.92 \\
\midrule
DASH-2, $B=5$
& 72.99
& 86.20
& 85.60
& 78.60
& 81.20
& 69.15
& 100.00
& 99.40
& 97.00
& 18.90
& 66.27
& 35.80
& 59.20
& 71.60 \\
DASH-2, $B=9$
& 86.88
& 95.40
& 98.40
& 94.60
& 96.85
& 92.50
& 100.00
& 99.80
& 99.80
& 69.52
& 71.60
& 47.80
& 70.87
& 92.28 \\
DASH-2, $B=18$
& 91.14
& 99.80
& 99.20
& 98.80
& 99.35
& 97.15
& 100.00
& 100.00
& 100.00
& 78.80
& 86.27
& 53.20
& 73.48
& 98.72 \\
\midrule
DASH-3, $B=4.875$
& 73.58
& 83.80
& 89.80
& 83.80
& 79.95
& 72.70
& 99.20
& 99.40
& 68.80
& 35.40
& 69.33
& 38.00
& 63.55
& 72.80 \\
DASH-3, $B=9.0$
& 87.63
& 97.00
& 98.60
& 96.40
& 97.85
& 90.95
& 100.00
& 99.60
& 100.00
& 72.74
& 76.47
& 46.60
& 71.15
& 91.80 \\
DASH-3, $B=17.75$
& 91.42
& 100.00
& 99.60
& 99.60
& 99.75
& 98.00
& 100.00
& 100.00
& 100.00
& 78.52
& 84.47
& 51.20
& 78.02
& 99.32 \\
\midrule
Jet-Nemotron-2B
& 83.86
& 97.00
& 91.60
& 94.40
& 96.65
& 92.10
& 100.00
& 100.00
& 100.00
& 61.54
& 70.73
& 44.00
& 71.12
& 71.08 \\
Jet-Nemotron-4B
& 83.71
& 99.00
& 93.20
& 94.20
& 95.15
& 98.10
& 100.00
& 100.00
& 99.40
& 60.96
& 78.00
& 53.00
& 70.48
& 46.80 \\
\bottomrule
\end{tabular}
}
\end{table*}

\FloatBarrier
\section{Searched Architectures and Allocation Trends}
\label{app:searched_architectures}

\subsection{Final Discrete Architectures}
\label{app:final_discrete_architectures}

We summarize and visualize the final discrete architectures used by
Interleave,
GA-S2,
and DASH.
Following Appendix~\ref{app:layer0_handling},
layer~0 is set to \textsc{Linear} in the final DASH architectures.
For tri-state architectures,
the realized budget is computed as
$n_{\rm FULL}+0.125 n_{\rm WINDOW}$.

\begin{table}[!htbp]
    \centering
    \small
    \caption{
    Summary of final operator counts and realized budgets.
    For binary architectures,
    the budget equals the number of \textsc{Full} layers.
    For tri-state DASH,
    the budget is computed as
    $n_{\rm FULL}+0.125n_{\rm WINDOW}$.
    }
    \label{tab:architecture_summary}
    \vspace{0.5em}
    \begin{tabular}{lcccc}
    \toprule
    Model
    & $n_{\rm FULL}$
    & $n_{\rm WINDOW}$
    & $n_{\rm LINEAR}$
    & Realized budget \\
    \midrule
    Interleave-5
    & 5 & 0 & 31 & 5 \\
    Interleave-9
    & 9 & 0 & 27 & 9 \\
    Interleave-18
    & 18 & 0 & 18 & 18 \\
    \midrule
    GA-S2-5
    & 5 & 0 & 31 & 5 \\
    GA-S2-9
    & 9 & 0 & 27 & 9 \\
    GA-S2-18
    & 18 & 0 & 18 & 18 \\
    \midrule
    DASH-2-5
    & 5 & 0 & 31 & 5 \\
    DASH-2-9
    & 9 & 0 & 27 & 9 \\
    DASH-2-18
    & 18 & 0 & 18 & 18 \\
    \midrule
    DASH-3-4.875
    & 4 & 7 & 25 & 4.875 \\
    DASH-3-9.0
    & 8 & 8 & 20 & 9.0 \\
    DASH-3-17.75
    & 17 & 6 & 13 & 17.75 \\
    \bottomrule
\end{tabular}
\end{table}

\begin{figure}[!htbp]
    \centering
    \includegraphics[
        width=\linewidth
    ]{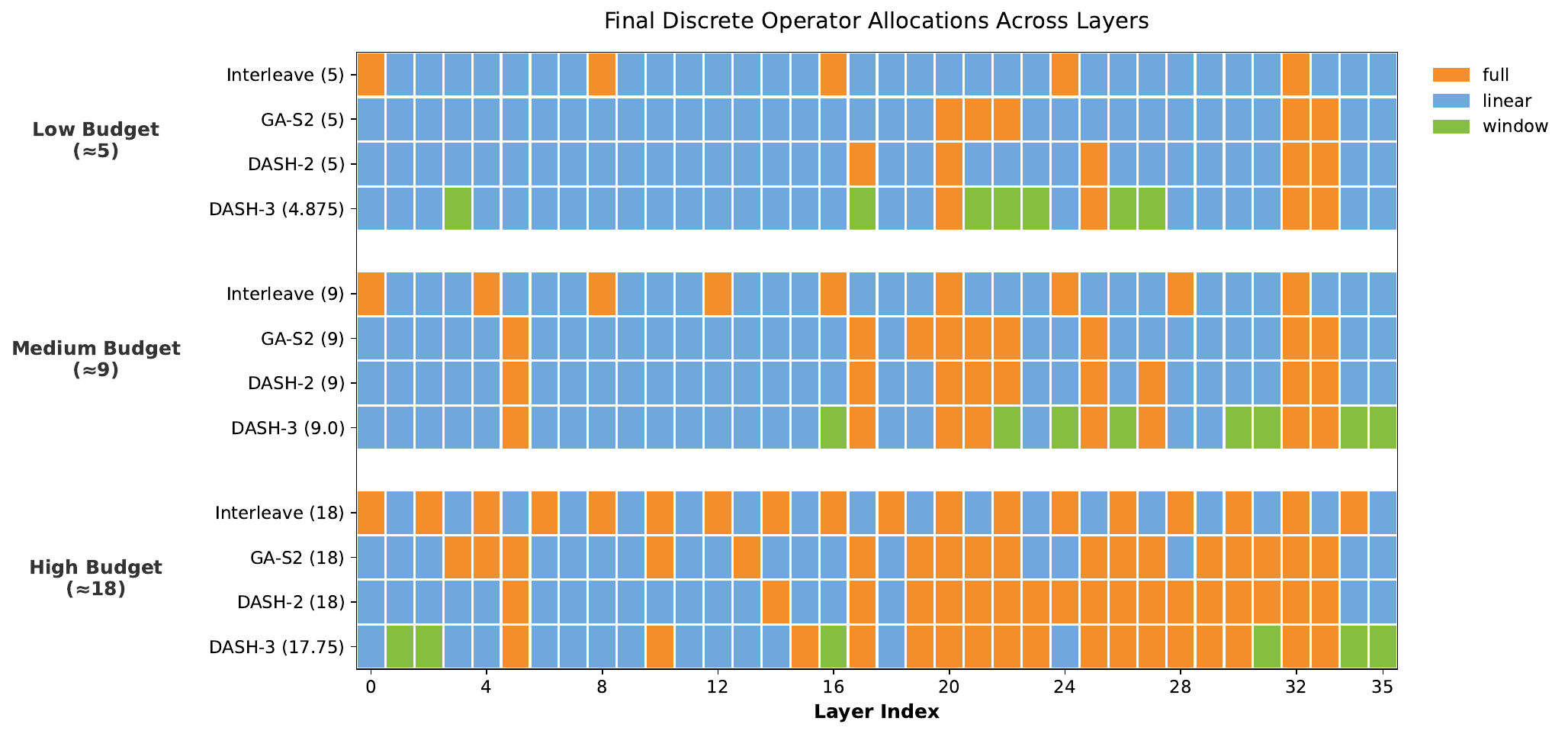}
    \caption{
    Final discrete operator allocations across layers.
    Each row corresponds to one evaluated architecture,
    and each column corresponds to a transformer layer.
    Colors denote \textsc{Full},
    \textsc{Linear},
    and \textsc{Window} operators.
    Rows are grouped by comparable budget levels.
}
\label{fig:final_operator_allocations}
\end{figure}

Table~\ref{tab:architecture_summary}
reports the operator counts and realized budgets of the evaluated
architectures,
while Figure~\ref{fig:final_operator_allocations}
visualizes their layer-wise operator placements.
Interleave follows fixed periodic patterns,
whereas GA-S2 and DASH place expensive operators more selectively.
DASH-3 additionally uses \textsc{Window} as an intermediate-cost
operator between \textsc{Linear} and \textsc{Full}.

\FloatBarrier
\subsection{Allocation Trends across Budgets}
\label{app:allocation_trends}

\begin{figure}[!htbp]
    \centering
    \includegraphics[
        width=\linewidth
    ]{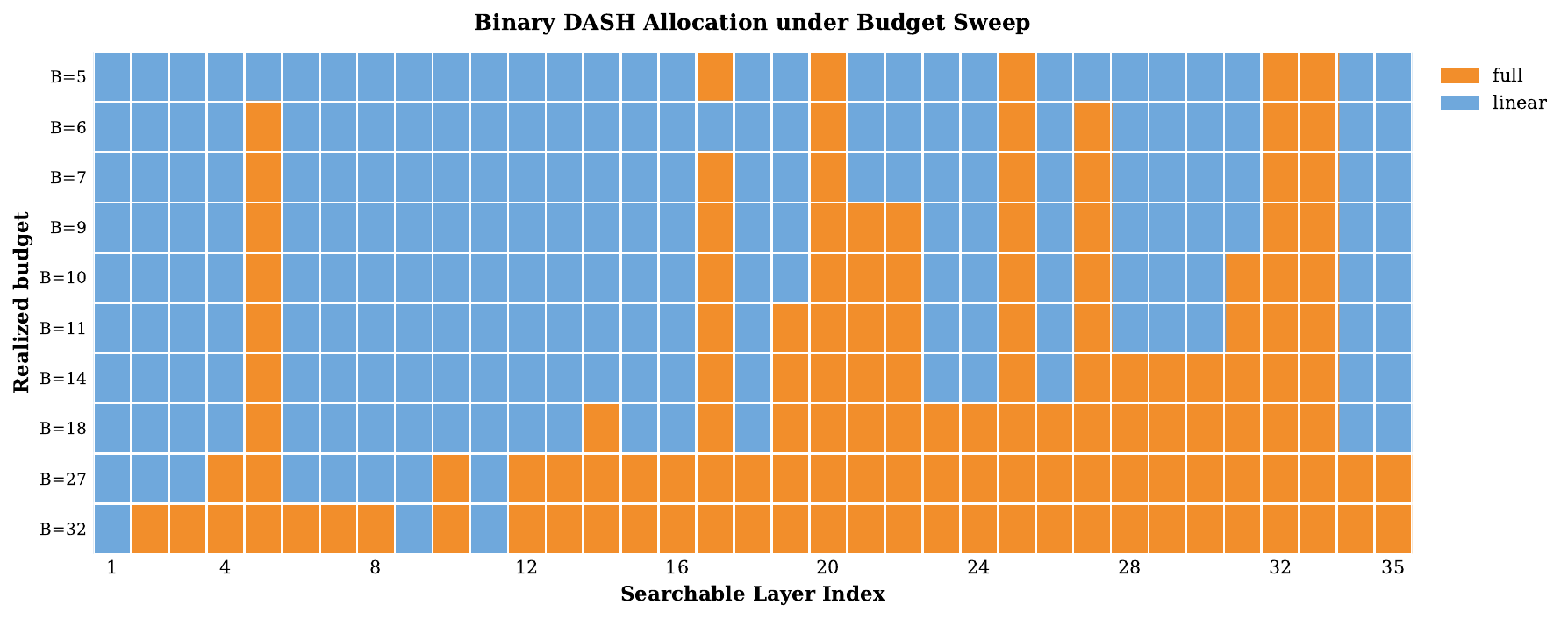}
    \caption{
    Binary DASH top-1 operator allocations across realized budgets.
    Each row corresponds to one searched architecture,
    sorted by realized budget,
    and each column corresponds to a searchable layer.
    Layer~0 is excluded because it is not searched in Stage~2.
    Colors indicate the top-1 discretized operator at each layer.
    }
\label{fig:binary_budget_sweep_allocation}
\end{figure}

We visualize how DASH allocation patterns 
vary with realized budget in
Figures~\ref{fig:binary_budget_sweep_allocation}
and~\ref{fig:tristate_budget_sweep_allocation}.
All runs use the Stage~2 search protocol in
Appendix~\ref{app:training_search_config}
and the layer-0 convention in
Appendix~\ref{app:layer0_handling}.

The visualizations show that increasing budget does not simply add
expensive operators uniformly across depth.
Instead,
DASH concentrates \textsc{Full} and \textsc{Window} choices on subsets
of layers,
while many layers remain \textsc{Linear}.
In the tri-state setting,
\textsc{Window} often serves as an intermediate allocation between
\textsc{Linear} and \textsc{Full},
supporting a finer-grained quality--efficiency trade-off.
We treat these plots as descriptive summaries of the searched
architectures,
rather than causal explanations of layer function.

\begin{figure}[!htbp]
    \centering
    \includegraphics[
        width=\linewidth
    ]{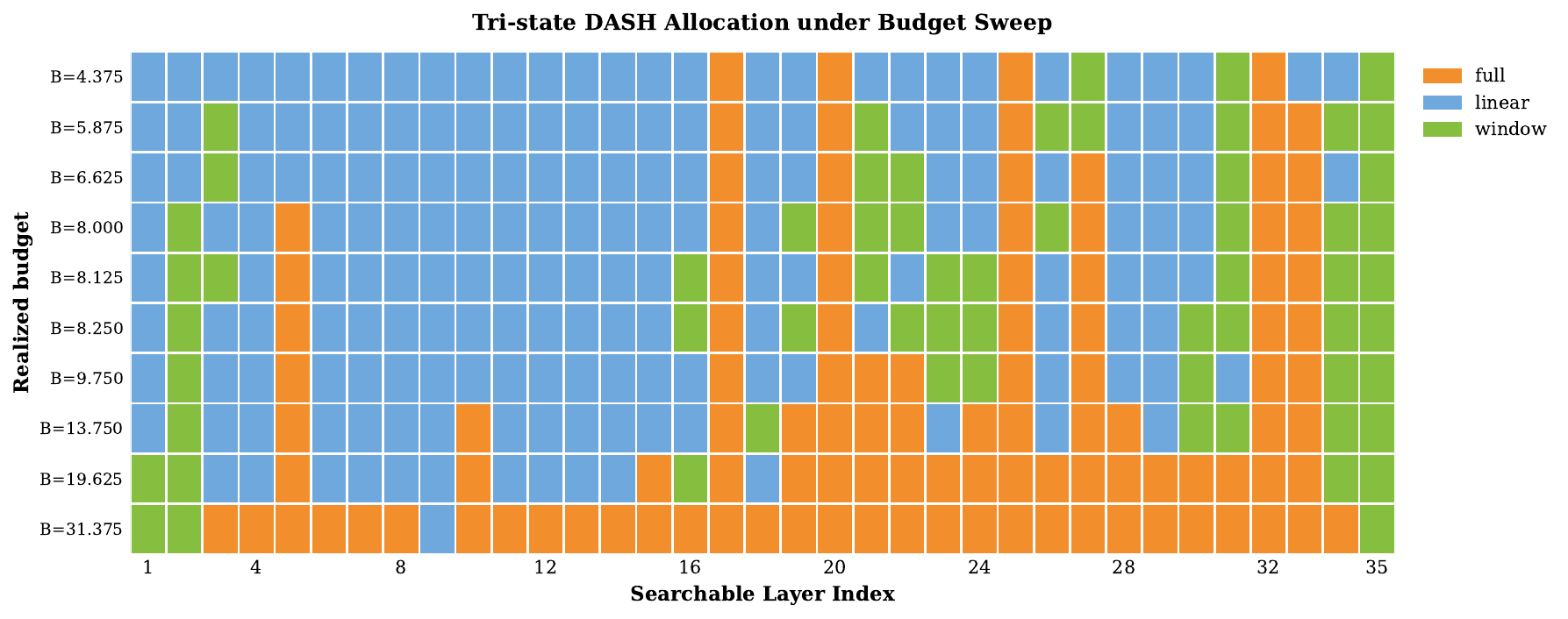}
    \caption{
    Tri-state DASH top-1 operator allocations across realized budgets.
    Each row corresponds to one searched architecture,
    sorted by realized budget,
    and each column corresponds to a searchable layer.
    Layer~0 is excluded because it is not searched in Stage~2.
    Colors indicate whether the top-1 operator is
    \textsc{Full},
    \textsc{Window},
    or \textsc{Linear}.
    }
    \label{fig:tristate_budget_sweep_allocation}
\end{figure}

\FloatBarrier
\section{Annealing Routing Visualization}
\label{app:annealing_routing_visualizations}

\begin{figure}[!htbp]
    \centering
    \includegraphics[width=0.85\linewidth]{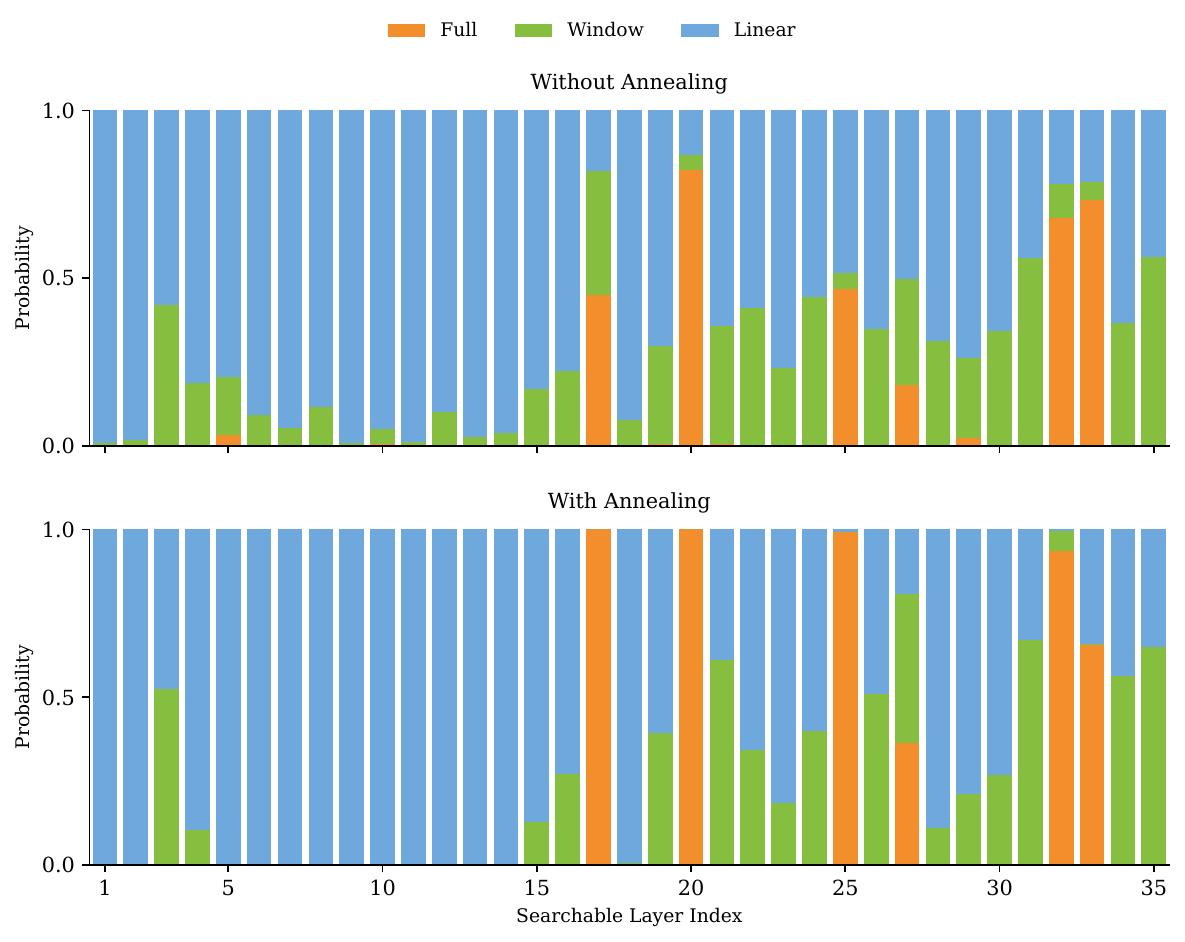}
    \caption{
Layer-wise routing distributions for a representative low-budget
tri-state pair with and without annealing.
Each stacked bar shows the final pre-discretization routing probabilities
over
\textsc{Full},
\textsc{Window},
and
\textsc{Linear}
for one searchable layer.
Annealing produces more concentrated distributions,
increasing top-1 confidence and reducing soft-to-hard discretization
mismatch.
    }
    \label{fig:anneal_alpha}
\end{figure}

We provide a representative visualization of the final soft routing
probabilities before hard discretization.
Figure~\ref{fig:anneal_alpha} shows a representative low-budget
tri-state pair with and without annealing,
using the same comparison protocol as the low-budget regime in
Table~\ref{tab:anneal_ablation}.
Together,
the routing diagnostics and probability visualization support our use of
annealing as a practical stabilizer for converting the differentiable
soft architecture into a discrete hybrid model.

\FloatBarrier
\section{Inference Efficiency Details}
\label{app:inference_protocol}

We report the measurement setup and raw numbers for the
inference-efficiency experiments in Section~\ref{sec:inference}.
All measurements are performed on a single NVIDIA RTX PRO 6000
Blackwell GPU with batch size 16.
We measure prefill latency,
decode latency,
and peak memory from 4K to 64K context length.
The setup is summarized in Table~\ref{tab:inference_setup},
and the raw prefill and decode results are reported in
Tables~\ref{tab:prefill_efficiency_raw}
and~\ref{tab:decode_efficiency_raw}.

\begin{table}[!htbp]
\centering
\small
\caption{
Inference measurement setup.
}
\label{tab:inference_setup}
\begin{tabular}{ll}
\toprule
Item & Value \\
\midrule
GPU memory & 95.6 GiB \\
Batch size & 16 \\
Precision & bfloat16 \\
Decode length & 256 tokens \\
Warmup & 3 warmup iterations \\
Measured iterations & 10 \\
\bottomrule
\end{tabular}
\end{table}

\begin{table}[!htbp]
\centering
\small
\caption{
Raw prefill latency and peak memory.
Each entry is reported as latency in seconds / peak memory in GiB.
}
\label{tab:prefill_efficiency_raw}
\begin{tabular}{lccccc}
\toprule
Model
& 4K
& 8K
& 16K
& 32K
& 64K \\
\midrule
Full-attention teacher
& 1.52 / 13.15
& 3.27 / 20.43
& 7.41 / 35.00
& 18.33 / 64.13
& OOM \\
DASH-3, $B=4.875$
& 1.34 / 13.07
& 2.86 / 18.35
& 5.87 / 26.73
& 12.24 / 43.48
& 26.58 / 78.63 \\
DASH-3, $B=9.0$
& 1.30 / 13.18
& 2.81 / 18.71
& 5.88 / 27.58
& 12.64 / 45.34
& 28.74 / 82.53 \\
DASH-3, $B=17.75$
& 1.27 / 13.51
& 2.82 / 19.61
& 6.12 / 29.61
& 13.96 / 49.61
& 34.61 / 91.36 \\
\bottomrule
\end{tabular}
\end{table}

\begin{table}[!htbp]
\centering
\small
\caption{
Raw decode latency and peak memory.
Each entry is reported as latency in ms/token / peak memory in GiB.
}
\label{tab:decode_efficiency_raw}
\begin{tabular}{lccccc}
\toprule
Model
& 4K
& 8K
& 16K
& 32K
& 64K \\
\midrule
Full-attention teacher
& 14.86 / 8.29
& 19.31 / 10.57
& 30.15 / 15.14
& 50.11 / 24.26
& OOM \\
DASH-3, $B=4.875$
& 22.26 / 7.88
& 22.16 / 8.19
& 22.45 / 8.82
& 22.29 / 10.07
& 22.42 / 12.66 \\
DASH-3, $B=9.0$
& 22.57 / 8.00
& 22.42 / 8.56
& 22.54 / 9.69
& 22.76 / 11.94
& 27.83 / 16.57 \\
DASH-3, $B=17.75$
& 22.07 / 8.37
& 21.93 / 9.49
& 22.01 / 11.74
& 28.12 / 16.25
& 46.66 / 25.43 \\
\bottomrule
\end{tabular}
\end{table}

\FloatBarrier
\section{Alternative Search Variants}
\label{app:alternative_search_variants}

During development,
we also explored two alternative search variants in the binary
\textsc{Full}/\textsc{Linear} setting.
The first is a bilevel-style variant that updates model weights during
search,
with the goal of allowing candidate operators to adapt jointly with
the architecture parameters.
The second is a straight-through variant that uses hard operator choices
in the forward pass,
with the goal of reducing the mismatch between soft search and hard
discretization.

We did not adopt these variants because they did not provide consistent
improvements in our pilot runs,
while increasing optimization complexity or reducing search stability.
The bilevel-style variant also entangles architecture selection with
search-time weight adaptation,
whereas our goal is to isolate the layer-allocation problem.
Accordingly,
DASH uses an architecture-only search stage:
model and operator weights are frozen,
only architecture logits are optimized,
and final weight adaptation is deferred to Stage~3.


\end{document}